\documentclass[journal]{IEEEtran}

\usepackage{array}
\usepackage{graphicx}
\usepackage{amsmath,amssymb,amsthm, amsfonts,bm}
\usepackage{gensymb}
\usepackage{resizegather}
\usepackage{tabularx} 

\usepackage{etoolbox}
\usepackage{floatrow}
\usepackage{algpseudocode}
\usepackage{algorithm}
\usepackage{comment}
\usepackage{tikz}
\usepackage{nicematrix}

\usepackage{dblfloatfix}

\newcommand*{\vect}[1]{\boldsymbol{#1}}

\newcommand*{\vectorize}[1]{\vectorizeoperator\{#1\}}
\DeclareMathOperator{\vectorizeoperator}{vec}

\begin{document}

\title{Reciprocal Visibility}

\author{Rakesh~John~Amala~Arokia~Nathan, Sigrid~Strand, Dmitriy~Shutin, and~Oliver~Bimber
        
\thanks{R.J.A.A. Nathan and O.Bimber are with the Department of Computer Science, Johannes Kepler University, 4040 Linz, Austria, e-mail: rakesh.amala\_arokia\_nathan@jku.at, oliver.bimber@jku.at. 
S. Strand and D. Shutin are with the Institute of Communications and Navigation Communications Systems, German Aerospace Center, 82234 Oberpfaffenhofen-Wessling, e-mail: sigrid.strand@dlr.de, dmitriy.shutin@dlr.de.
}


\thanks{}}

\markboth{}%
{Nathan \MakeLowercase{\textit{et al.}}: Reciprocal Visibility}

\maketitle

\begin{abstract}
We propose a guidance strategy to optimize real-time synthetic aperture sampling for occlusion removal with drones by pre-scanned point-cloud data. Depth information can be used to compute visibility of points on the ground for individual drone positions in the air. Inspired by Helmholtz reciprocity, we introduce reciprocal visibility to determine the dual situation - the visibility of potential sampling position in the air from given points of interest on the ground. The resulting visibility map encodes which point on the ground is visible by which magnitude from any position in the air. Based on such a map, we demonstrate a first greedy sampling optimization.   
\end{abstract}

\IEEEpeerreviewmaketitle

\section{Introduction}\label{Sec:Introduction}
For many applications that apply aerial imaging, occlusion caused by dense vegetation (such as forest) represents a fundamental problem. Examples are search and rescue, wildfire detection, wildlife observation, surveillance, forestry, agriculture, and archaeology. Synthetic aperture imaging approaches, such as Airborne Optical Sectioning (AOS) \cite{Kurmi19}, remove occlusion in aerial images computationally in real-time.  
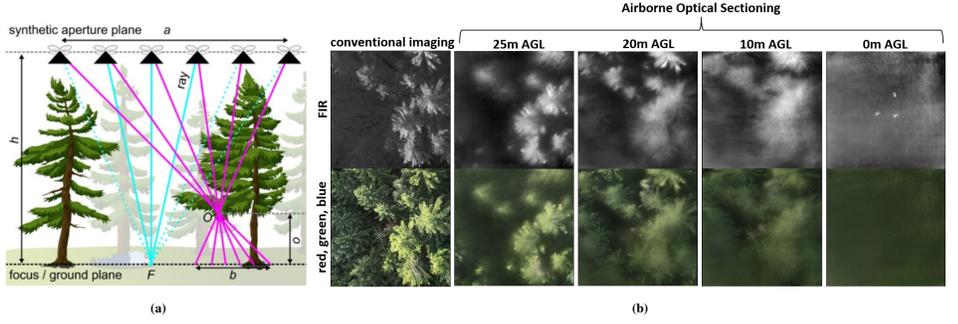
\begin{figure*}[!t]

    \begin{tikzpicture}
    \useasboundingbox (0,0) rectangle (12.5,0);
    \node [anchor=center,text width=0.5cm] (eq_col_a) at (5.3,-4.1) 
    {{\tiny\textbf{(a)}}};
    \node [anchor=center,text width=0.5cm] (eq_col_a) at (11.7,-4.1) 
    {{\tiny\textbf{(b)}}};
    \end{tikzpicture}
    \floatbox[{\capbeside\thisfloatsetup{capbesideposition={left,top},capbesidewidth=5.5cm}}]{figure}[12.5cm]
    {\caption{Synthetic aperture imaging and integration principle (a). Examples of conventional aerial images and AOS integral images of forest at different synthetic focal distances and captured at various spectral ranges (visible red/green/blue and far-infrared, FIR)(b). This examples illustrates a search and rescue use-cases where in particular thermal measurements are relevant. People on the ground become visible in the FIR channel after occlusion removal (i.e., by focusing computationally on the ground). This data has been recorded with a single drone in a 30m x 30m  waypoint grid and at an altitude of 35m above ground level (AGL) \cite{Schedl20b}.}\label{fig:test}}
    {\includegraphics[width=1.0\linewidth]{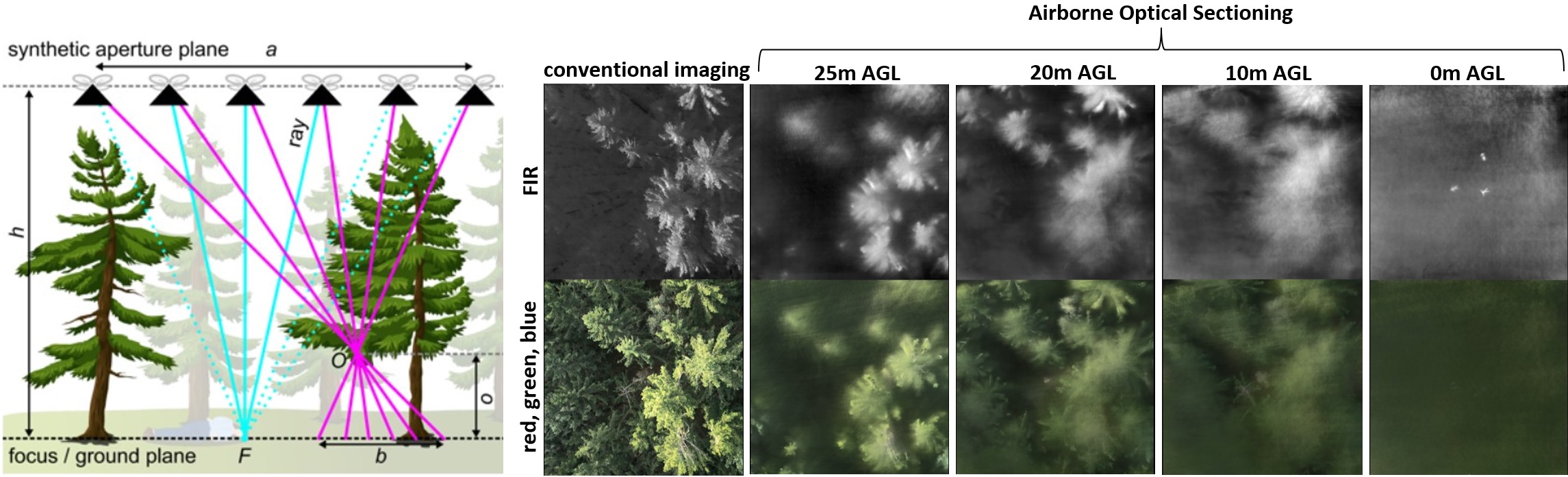}}
    \label{FIG:AOS}
\end{figure*}
As illustrated in Fig. \ref{FIG:AOS}a, AOS registers and integrates (i.e., averages) multiple images captured with conventional camera optics at different drone positions to an integral image that mimics a wide (several meters) synthetic camera aperture. The area $a$ in which these images are sampled is the size of the synthetic aperture. Registration and integration depend on the camera poses and on a defined focal surface $F$ (e.g., a given plane or a registered digital elevation map of the ground surface at distance $h$). The computed integral images (Fig. \ref{FIG:AOS}b) have an extremely shallow depth of field. This causes all targets located on the focal surface at distance $F$ to appear sharp and unoccluded while the blur signal of the occluders at distance $o$ from $F$ (i.e., not on the focal surface) is spread widely ($b$) over the integral image (which suppresses their contribution). The main advantages of AOS over alternatives, such as light detecting and ranging (LiDAR) or synthetic aperture radar (SAR), are high spatial resolution, real-time processing capabilities, and wavelength independents. Since it is applicable to images captured in all spectral ranges and processing is in the range of milliseconds, AOS has been demonstrated for use-cases, such as search and rescue \cite{Schedl20b}\cite{Schedl21}, wildlife observation \cite{Schedl20a}, wildfire detection \cite{Youssef23}, and archaeology \cite{Kurmi19}.

Not only raw images can be integrated with AOS, but also pre-processed ones. Anomaly detectors (i.e., the widely used Reed-Xiaoli (RX) detector \cite{reed1990adaptive}), for instance, mark pixels of abnormal appearance, such as color, temperature, or motion behavior. The binary masks that result when anomaly detection is applied to an aerial image of occluding forest can be considered as a simple visibility mask. The majority of pixels represent the common --and therefore normal-- forest background (mapped to 0), while appearances of unoccluded targets on the ground is usually abnormal (mapped to 1). Integrating such anomaly masks instead of raw images with AOS, leads to a 2D integral visibility map that accumulates visibility from all drone poses within the synthetic aperture, as explained in more detail in \cite{Nathan23a}.    

A main challenge of synthetic aperture imaging approaches, such as AOS, is optimal sampling (i.e., the question where to position the drones within the synthetic aperture to achieve highest integrated visibility with the smallest number of samples). Several approaches, ranging from classical predefined waypoint flights (1D lines and 2D grids) with single drones \cite{Kurmi19}\cite{Schedl20a}, over parallel sampling with camera arrays \cite{Nathan22}, to particle swarm optimization with drones swarms \cite{Nathan23b} have been investigated. The latter supports adapted sampling with respect to local forest density or target appearance and outperforms any blind sampling strategies, such as fixed waypoint flights.

In this article, we propose to guide real-time AOS sampling by pre-captured point-cloud data that results from LiDAR scans. Although such data is usually not suitable for real-time occlusion removal (too low resolution, does not support required wavelengths) it can be beneficial for  optimized sampling. Given a (possibly sparse) point cloud of forest, the contained depth information can be used to directly compute single binary visibility masks for all drone perspectives by filtering out 3D projected points that are not near or not on the ground surface. 

In this work, we want to introduce the principle of \textit{reciprocal visibility} (RV) in Sect. \ref{Sec:Theory}, explain how this principle can be practically implemented in Sect.  \ref{Sec:Implementation}, and show how RV can be applied to optimize AOS sampling in Sect. \ref{Sec:Sampling}. For our discussion, we apply the same procedural forest model as in \cite{Nathan23b} to simulate 3D data, which would normally result from LiDAR scans. We show the potential advantage of RV-guided sampling over previous unguided sampling, such as \cite{Nathan23b}.  
               
\section{Theory} \label{Sec:Theory}
Let us consider the synthetic aperture imaging and integration process as described previously (see also Fig.~\ref{FIG:AOS}a). Let us further consider $N$ visibility masks $\vect{V}_n$, $n=1,\ldots,N$, at the resolution $M=R_v\times R_v$ pixels per mask, for all drone positions over the synthetic aperture plane, as explained above.
The resulting and vectorized $R_v\times R_v$ pixel integral visibility map $\vect{i}$ can then be formed by a convex combination of the collected single visibility masks as follows:
\begin{equation}\label{Eq:ForwardVisibility}
    \vect{i}= \frac{1}{Z}\sum_{n=1}^N \vect{v}_n p_n=\vect{V}\vect{p},
\end{equation}
where $\vect{p}=[p_1,\ldots,p_N]^\top$ represents the synthetic aperture by a \emph{binary} vector that indicates which visibility maps are included into $\vect{i}$, $Z=\|\vect{p}\|^2$ is a normalization constant, and $\vect{V}=[\vect{v}_1,\ldots,\vect{v}_N]$ is a so-called visibility matrix. 
Its columns are defined as vectorized representations of the registered visibility maps $\vect{v}_n=\vectorize{\vect{V}_n}$.
Note that $\vect{V}$ is a $M\times N$ matrix; also, w.l.o.g. we absorb the scalar $Z$ into $\vect{V}$ to simplify further notation.

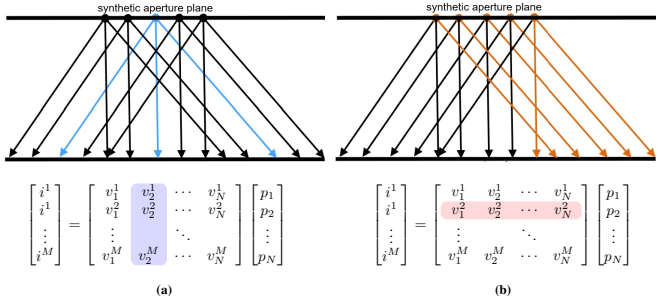
\begin{figure}[!h]
    \centering
    \begin{tikzpicture}
    \node[anchor=south west,inner sep=0] (column) at (-0.4,1.5) {\resizebox{0.48\linewidth}{!}{\includegraphics{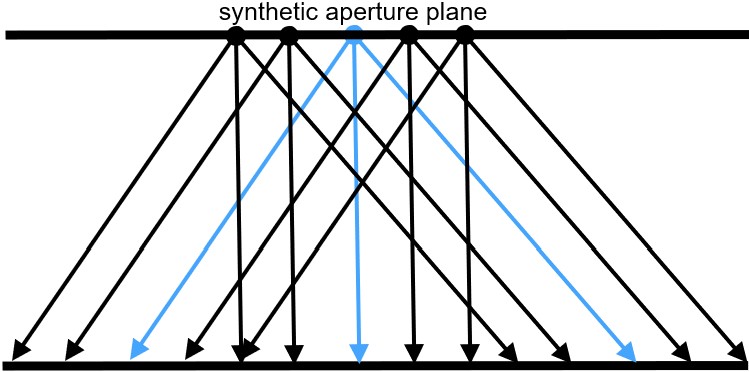}}}; 
    \node[anchor=south west,inner sep=0] (row) at (4,1.5) {\resizebox{0.48\linewidth}{!}{\includegraphics{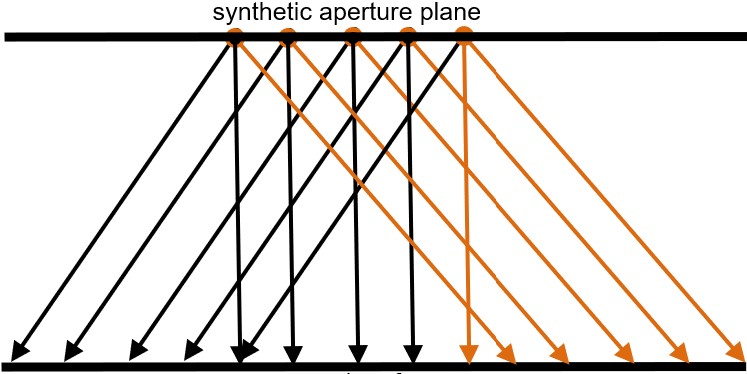}}};

    \node [anchor=south west,text width=6cm] (eq_col) at (-0.3,0) {
        \begin{minipage}{\textwidth}
        \resizebox{0.6\textwidth}{!}{
                $
                \begin{bmatrix}
                i^1\\
                i^1\\
                \vdots\\
                i^M
                \end{bmatrix} =
                \begin{bNiceMatrix}[margin]
                v^1_1  & \Block[fill=blue!15,rounded-corners]{4-1}{} v^1_2 & \cdots & v^1_N\\
                v^2_1  & v^2_2  & \cdots & v^2_N\\
                \vdots &        & \ddots       &      \\
                v^M_1 & v^M_2 & \cdots & v^M_N
                \end{bNiceMatrix}
                \begin{bmatrix}
                p_1\\
                p_2\\
                \vdots\\
                p_N
                \end{bmatrix}
                $  
                }
        \end{minipage}
    };
        \node [anchor=south west,text width=6cm] (eq_col_a) at (1.5,-0.4) {{\tiny\textbf{(a)}}};
    
    \node [anchor=south west,text width=6cm] (eq_row) at (4.3,0) {
        \begin{minipage}{\textwidth}
        \resizebox{0.6\textwidth}{!}{
                $
                \begin{bmatrix}
                i^1\\
                i^1\\
                \vdots\\
                i^M
                \end{bmatrix} =
                \begin{bNiceMatrix}[margin]
                v^1_1  & v^1_2  & \cdots & v^1_N\\
                \Block[fill=red!15,rounded-corners]{1-4}{}
                v^2_1  & v^2_2  & \cdots & v^2_N\\
                \vdots &        & \ddots       &      \\
                v^M_1 & v^M_2 & \cdots & v^M_N
                \end{bNiceMatrix} 
                \begin{bmatrix}
                p_1\\
                p_2\\
                \vdots\\
                p_N
                \end{bmatrix}
                $  
                }
        \end{minipage}
    };
    \node [anchor=south west,text width=6cm] (eq_col_b) at (6,-0.4) {{\tiny\textbf{(b)}}};

  \end{tikzpicture}
    \vspace{-0.6cm}
    \caption{%
    2D illustration of the 4D visibility field in $\vect{V}$. While same columns of $\vect{V}$ represent same positions on the synthetic aperture plane (a), same rows of $\vect{V}$ represent same directions (b).%
    }
    \label{FIG:pos_dir_RV}
\end{figure}

The structure of $\vect{V}$ bears the key to understanding the reciprocal visibility principle.
The element $v_n^m=[\vect{V}]_{m,n}$, i.e., an entry of $\vect{V}$ at the $n$th column and the $m$th row, corresponds to the visibility from the $n$th position on the synthetic aperture plane at the direction $m$.
Thus, while a single column $\vect{v}_n$ of $\vect{V}$ encodes the visibility at a fixed aperture plane position $n$, a single row encodes visibility along a single direction $m$ over different aperture positions (see also Fig.\ref{FIG:pos_dir_RV}). 
Note that in this way the 2D structure of $\vect{V}$ effectively encodes an actual 4D visibility field.

Now, given an integral visibility map $\vect{i}$, we can identify an optimal sampling pattern $\vect{p}$ on the synthetic aperture plane e.g., as a solution to an optimization problem $\min_{\vect{p}} \|\vect{i}-\vect{V}\vect{p}\|$ with respect to some preferred norm $\|\cdot\|$. 
Unfortunately, such an optimization problem is ill-posed and computationally very challenging for realistic processing times and memory space for typical values of $M$ and $N$.

A computationally more attractive approach in this respect relies on the principle of Helmholz reciprocity \cite{Stokes49} \cite{Helmholtz56} which has been previously applied in computational photography to 2D images and a projector-camera pair \cite{Sen05}.
This principle states that the transport of light which passes through a passive medium is dual with respect to its travel direction: an emitter and a receiver can be swapped, while ensuring the same light-transport. 
Helmholz reciprocity applies to all passive media, and covers reflection, refraction, absorption, as well as occlusion. 
In our setting this principle can be applied as follows: 

Assume a reverse scenario, where the synthetic aperture plane is now located on the ground, and the drone camera is looking upwards. 
In this case \eqref{Eq:ForwardVisibility} can be transformed as 
\begin{equation}\label{Eq:ReciprocalVisibility}
    \vect{i}_{\uparrow}= \vect{V}^\top\vect{p}_{\uparrow},
\end{equation}
where we used a subscript $(\cdot)_\uparrow$ to explicitly indicate that here the integrated visibility map is of the ``sky''  obtained by ``looking'' upwards from the synthetic aperture plane on the ground.
Thus, $\vect{p}_{\uparrow}$ is a binary aperture selection vector again, and $\vect{i}_{\uparrow}$ is the ``sky'' integral visibility.\footnote{Again here an integral visibility $\vect{i}_{\uparrow}$ is a convex combination of columns in $\vect{V}^\top$, implying a normalization constant $1/Z=1/\|\vect{p}_{\uparrow}\|^2$.}
Thus, in \eqref{Eq:ForwardVisibility} $\vect{V}$ describes the visibility field in top-down direction.
Its transpose in \eqref{Eq:ReciprocalVisibility} describes the dual visibility field in bottom-up direction. 

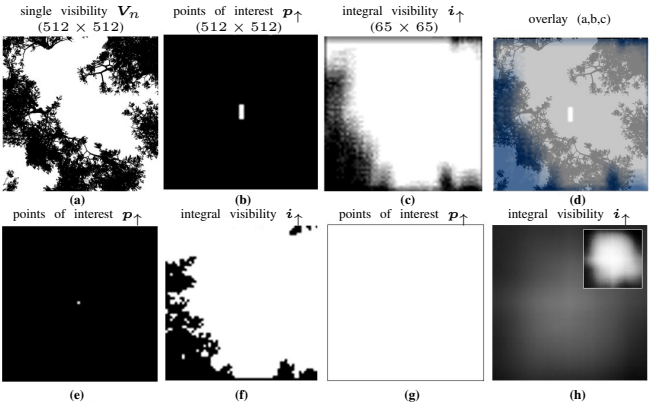
\begin{figure}[!h]
    \centering
    \begin{tikzpicture}
    \node[anchor=south west,inner sep=0] (column) at (-0.4,0) 
    {\resizebox{0.98\linewidth}{!}{\includegraphics{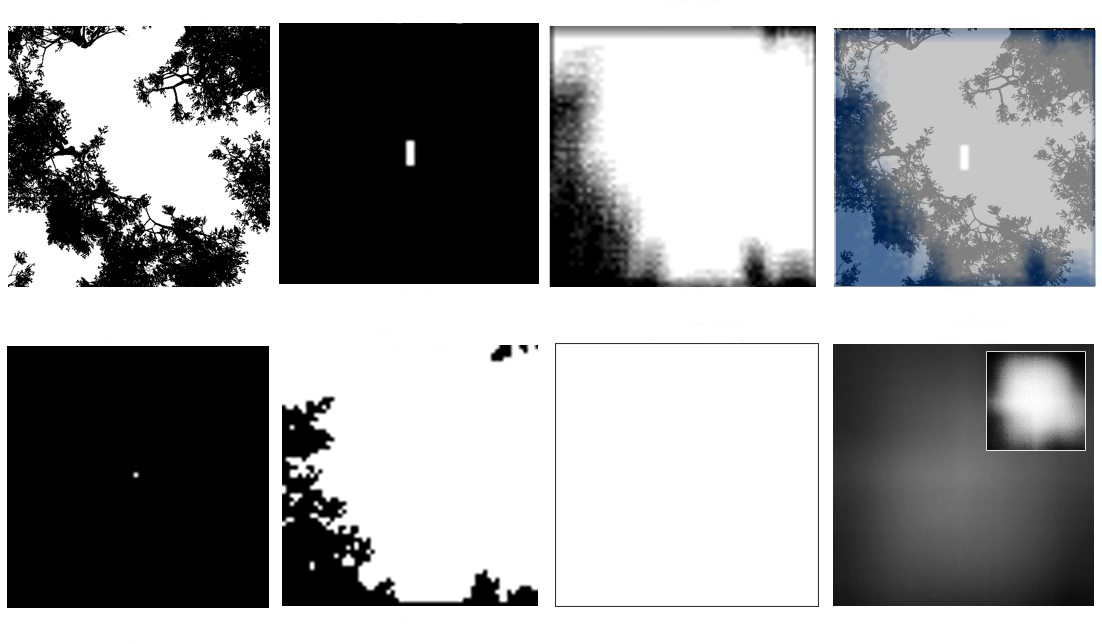}}}; 
    \node [anchor=center,text width=2cm] (eq_col_a) at (0.7,5.1) {
        \begin{minipage}{\textwidth}
        \tiny\centering
        single visibility $\vect{V}_n$\\
         ($512\times 512$)
        \end{minipage}
    };
    \node [anchor=center,text width=0.5cm] (eq_col_a) at (0.8,2.7) 
    {{\tiny\textbf{(a)}}};
    
    \node [anchor=center,text width=2cm] (eq_col_a) at (2.8,5.1) {
        \begin{minipage}{\textwidth}
        \tiny\centering
        points of interest $\vect{p}_{\uparrow}$\\
        ($512\times 512$)
        \end{minipage}
    };
    \node [anchor=center,text width=0.5cm] (eq_col_a) at (3,2.7) 
    {{\tiny\textbf{(b)}}};
    
    \node [anchor=center,text width=2cm] (eq_col_a) at (5,5.1) {
        \begin{minipage}{\textwidth}
        \tiny\centering
        integral visibility $\vect{i}_{\uparrow}$\\
       ($65\times 65$)
        \end{minipage}
    };
    \node [anchor=center,text width=0.5cm] (eq_col_a) at (5.2,2.7) 
    {{\tiny\textbf{(c)}}};

    \node [anchor=center,text width=2cm] (eq_col_a) at (7.2,5.1) {
        \begin{minipage}{\textwidth}
        \tiny\centering
        overlay (a,b,c)
        \end{minipage}
    };
    \node [anchor=center,text width=0.5cm] (eq_col_a) at (7.45,2.7) 
    {{\tiny\textbf{(d)}}};

    \node [anchor=center,text width=2cm] (eq_col_a) at (0.8,2.47) {
        \begin{minipage}{\textwidth}
        \tiny
        points of interest $\vect{p}_{\uparrow}$
        \end{minipage}
    };
    \node [anchor=center,text width=0.5cm] (eq_col_a) at (0.8,0.1) 
    {{\tiny\textbf{(e)}}};
    
    \node [anchor=center,text width=2cm] (eq_col_a) at (2.85,2.47) {
        \begin{minipage}{\textwidth}
        \tiny\centering
        integral visibility $\vect{i}_{\uparrow}$
        \end{minipage}
    };
    \node [anchor=center,text width=0.5cm] (eq_col_a) at (3,0.1) 
    {{\tiny\textbf{(f)}}};
    
    \node [anchor=center,text width=2cm] (eq_col_a) at (5,2.47) {
        \begin{minipage}{\textwidth}
        \tiny\centering
        points of interest $\vect{p}_{\uparrow}$
        \end{minipage}
    };
    \node [anchor=center,text width=0.5cm] (eq_col_a) at (5.25,0.1) 
    {{\tiny\textbf{(g)}}};

    \node [anchor=center,text width=2cm] (eq_col_a) at (7.2,2.47) {
        \begin{minipage}{\textwidth}
        \tiny\centering
            integral visibility $\vect{i}_{\uparrow}$
        \end{minipage}
    };
    \node [anchor=center,text width=0.5cm] (eq_col_a) at (7.45,0.1) 
    {{\tiny\textbf{(h)}}};

  \end{tikzpicture}
    \vspace{-0.3cm}
    \caption{%
    2D visualizations of components in \eqref{Eq:ReciprocalVisibility} using procedural forest simulation. Single visibility masks $\vect{V_n}$ (a) indicate visibility (binary 1) and occlusion (binary 0) as seen top-down from the synthetic aperture plane to the ground. For selected points of interest on the ground $\vect{p}_{\uparrow}$ (b,e,g), the integrated visibility maps $\vect{i}_{\uparrow}$ (c,f,h) indicate the visibility of the selected ground points from all positions at the synthetic aperture plane (higher values correspond to higher visibility). The inlay in (h) is a contrast enhanced version for better visibility. An overlay of (a,b,c) illustrates that the resulting integrated visibility spatially aligns correctly with the occluder situation of the forest (here, darker blue indicates lower visibility of the selected ground points) -- keeping tree heights (20m on average in this example) and altitude of the synthetic aperture plane (35m above ground level in this example) in mind. Simulated density: 100$trees/ha$ (birch) on a 32x32$m^2$ area.%
    }
    
    \label{FIG:RV}
\end{figure}

To better understand this concept in a relevant application setting consider a realization of \eqref{Eq:ForwardVisibility} and \eqref{Eq:ReciprocalVisibility} in Fig.~\ref{FIG:RV}.

Here we show a single visibility mask $\vect{V}_n$ with $M=512\times 512$ pixels of a simulated procedural birch forest as seen top-down, from the center perspective of the synthetic aperture at $35$m altitude above ground level, covering a 32x32$m^2$ area with a 50$^{\circ}$ field of view camera (Fig. \ref{FIG:RV}a). 
Assume also that a total of $N=65^2=4.225$ such visibility maps is available (at $65\times 65$ synthetic aperture grid points), and consider a binary pattern of ground points of interest $\vect{p}_{\uparrow}$ of resolution $512\times 512$ (Fig. \ref{FIG:RV}b).
Naturally, the integral visibility $\vect{i}_{\uparrow}$ (Fig.~\ref{FIG:RV}c) will have $N$ pixels. 
If $\vect{p}_{\uparrow}$ contains only a single point of interest on the ground (see Fig.~\ref{FIG:RV}e), then, according to \eqref{Eq:ReciprocalVisibility},  $\vect{i}_{\uparrow}$ is a pinhole-view of the ``sky'', as seen from this point upwards (see Fig.~\ref{FIG:RV}f).
In case $\vect{p}_{\uparrow}=\vect{1}$, i.e., all ground points are of interest (Fig. \ref{FIG:RV}g), the resulting $\vect{i}_{\uparrow}$ is shown in Fig. \ref{FIG:RV}h. 
Higher values indicate higher visibility of the selected ground points.    
Note that since typically $\vect{V}$ is a tall matrix, i.e., $M\le N$, in reciprocal visibility an integral visibility map $\vect{i}_{\uparrow}$ will typically have a lower resolution. 

Unfortunately, the complexity of the approach presented above is clearly not practical. Even in the considered example with $N=4.225$, the visibility matrix $\vect{V}$ will consist of $\sim10^9$ coefficients: the corresponding sampling time would be massive, yet it will lead to a rather low resolution of the resulting integrated visibility. 

In the following we discuss how this theoretical principle can be implemented efficiently in practice and how the resolution of the integral visibility can be improved. 
\section{Practical Reciprocal Visibility Implementation}\label{Sec:Implementation}
Let us restate our objective\--- we aim for an integral visibility map that would tell us which points on the ground will be visible at different synthetic aperture positions in the air. 
This information can then be re-used to optimally navigate camera-equipped drones for overall occlusion removal.

\begin{figure}[!h]
    \centering
    \begin{tikzpicture}
    \node[anchor=south west,inner sep=0] (column) at (-0.4,0) 
    {\resizebox{0.98\linewidth}{!}{\includegraphics[width=0.95\linewidth]{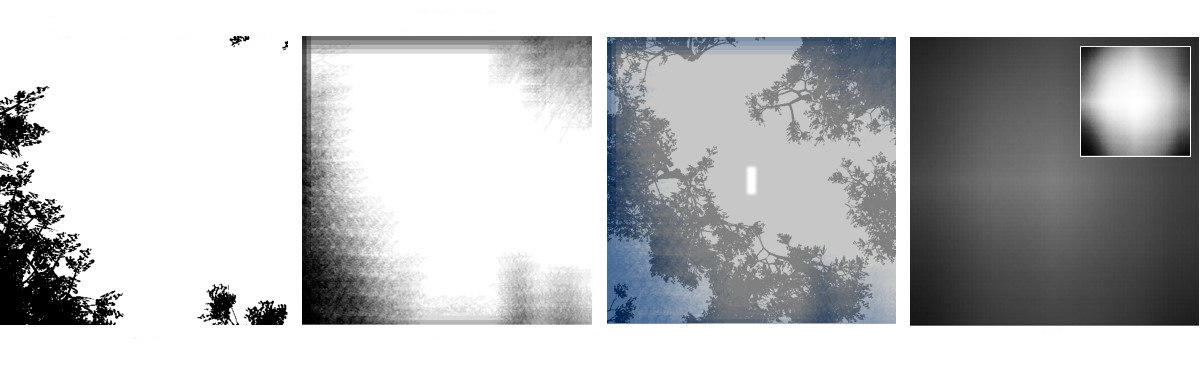}}}; 
    \node [anchor=center,text width=2cm] (eq_col_a) at (0.65,2.59) {
        \begin{minipage}{\textwidth}
        \tiny
        \centering
        single column of $\vect{U}_{\uparrow}$\\
        $(512\times512)$
        \end{minipage}
    };
    \node [anchor=center,text width=0.5cm] (eq_col_a) at (0.8,0.1) 
    {{\tiny\textbf{(a)}}};
    
    \node [anchor=center,text width=2cm] (eq_col_a) at (2.85,2.59) {
        \begin{minipage}{\textwidth}
        \tiny\centering
        integral visibility $\vect{i}_{\uparrow}$\\
        $(512\times512)$
        \end{minipage}
    };
    \node [anchor=center,text width=0.5cm] (eq_col_a) at (3.1,0.1) 
    {{\tiny\textbf{(b)}}};
    
    \node [anchor=center,text width=2cm] (eq_col_a) at (5,2.59) {
        \begin{minipage}{\textwidth}
        \tiny\centering
        overlay (a,b,Fig.\ref{FIG:RV}b)
        \end{minipage}
    };
    \node [anchor=center,text width=0.5cm] (eq_col_a) at (5.25,0.1) 
    {{\tiny\textbf{(c)}}};

    \node [anchor=center,text width=2cm] (eq_col_a) at (7.2,2.59) {
        \begin{minipage}{\textwidth}
        \tiny\centering
            integral visibility $\vect{i}_{\uparrow}$
            \\
        $(512\times512)$
        \end{minipage}
    };
    \node [anchor=center,text width=0.5cm] (eq_col_a) at (7.4,0.1) 
    {{\tiny\textbf{(d)}}};

  \end{tikzpicture}
    \vspace{-0.8cm}
    \caption{%
    A single bottom-up visibility mask (one column of $\vect{U}_{\uparrow}$) computed from a ground point of interest -- center ground position in this example -- (a) is the high-resolution counter-part of the integrated visibility map $i_{\uparrow}$ for the sample point in \eqref{Eq:ReciprocalVisibility} - see Fig. \ref{FIG:RV}f. Corresponding high-resolution integrated visibility maps (b for Fig.~\ref{FIG:RV}c, d for Fig.~\ref{FIG:RV}h) and the overlay visualization (d) that corresponds to Fig.~\ref{FIG:RV}d.  Here, the integrated visibility map in (b,c) has been computed for only 21 ground points of interest that approximates the shape of our rectangular pattern shown in (c). Simulated density: 100$trees/ha$ on 32x32$m^2$.%
    }
    \label{FIG:bottom-upRV}
\end{figure}
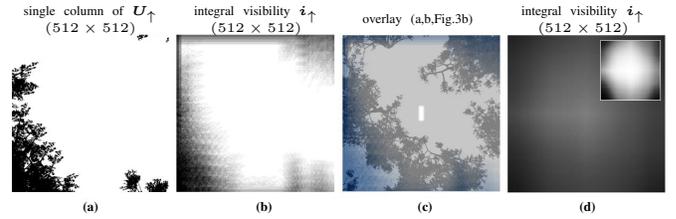

To this end, let us now assume that we are given (or we can design) a pattern $\vect{p}_{\uparrow}$ that represents points of interest on the ground.
The reverse imaging would imply an availability of the corresponding visibility matrix describing visibility from bottom upwards, following \eqref{Eq:ReciprocalVisibility}.
The latter is unfortunately physically unavailable.   
Instead, consider a scanning drone, equipped with e.g., a LIDAR that can generate a (possibly sparse) point cloud representation of the forest. 
These collected scans, although sparse, are in fact top-down ``proxies'' of $\vect{V}_n$ used in \eqref{Eq:ForwardVisibility}. 
Although point clouds cannot be used directly, they can be 3D projected to compute a high-resolution, e.g., $512\times 512$, equivalent of a bottom-up visibility map $\vect{U}_{\uparrow}\approx \vect{V}^\top$, i.e., from positions on the ground surface upwards.
This process is illustrated in Fig. \ref{FIG:bottom-upRV}. 
Here we consider $\vect{p}_{\uparrow}$ as shown in Fig.~\ref{FIG:RV}b, i.e., a rectagular area of $21$ points on the ground.
A visibility image corresponding to one of these points, i.e., to a single column of $\vect{U}_{\uparrow}$, is exemplary shown in Fig.~\ref{FIG:bottom-upRV}a. 
Again, we use a simulated data here, but in practice a computation reconstruction of the visibility image from LiDAR data should be used. 
Note that $\vect{U}_{\uparrow}$ is a tall matrix, leading to a high-resolution integral image $\vect{i}_{\uparrow}=\vect{U}_{\uparrow}\vect{p}_{\uparrow}$ in Fig.~\ref{FIG:bottom-upRV}b.
In case $\vect{p}_{\uparrow}=\vect{1}$, i.e., all points on the ground are considered, we obtain in Fig.~\ref{FIG:bottom-upRV}d a high-resolution version of integral visibility Fig.~\ref{FIG:RV}h.

One remaining problem of this approach is that although practical, the computed integral visibility $\vect{i}_{\uparrow}$ completely obscures which pixel of $\vect{p}_{\uparrow}$ is visible.
In other words, we obtain an integrated visibility of all points on the ground that are selected in $\vect{p}_{\uparrow}$.  
\begin{figure}[!h]
    \centering
    \begin{tikzpicture}
    \useasboundingbox (0,0) rectangle (12.5,0);
   
    \node [anchor=center,text width=0.5cm] (eq_col_a) at (2.3,-4.0) 
    {{\tiny\textbf{(a)}}};
    \node [anchor=center,text width=0.5cm] (eq_col_a) at (6.8,-4.0) 
    {{\tiny\textbf{(b)}}};
    \end{tikzpicture}
    \vspace{0.2cm}
    \includegraphics[width=0.95\linewidth]{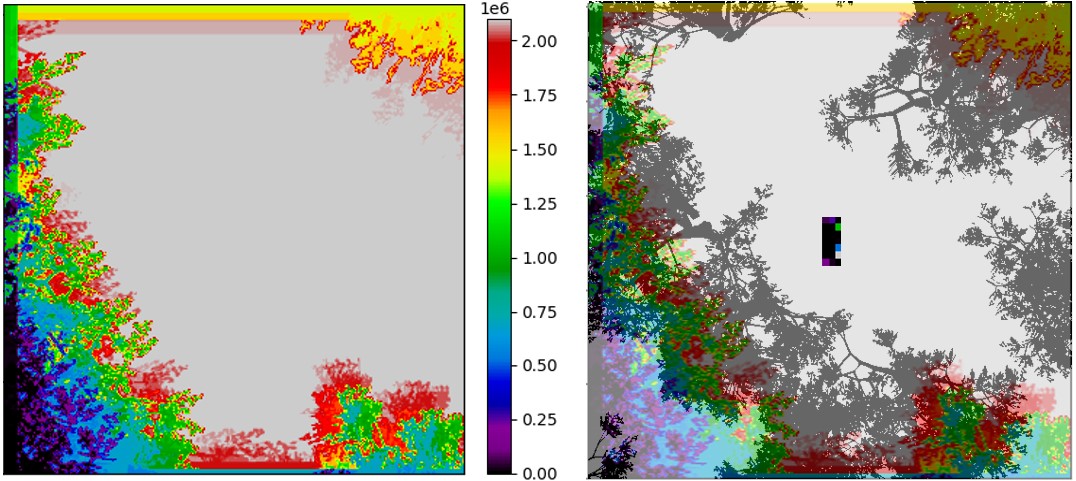}
    \vspace{-0.4em}\caption{%
    Coded integrated visibility map $\vect{i}_{\uparrow}$ for the example shown in Fig. \ref{FIG:bottom-upRV}b (a) and the overlay (b) that corresponds to Fig. \ref{FIG:bottom-upRV}d. In this case, unique binary codes for $K=21$ ground points of interest (individual codes are now visualized in our target area) are integrated. The possible $2^{21}=2.097.152$ different integrated combinations in $\vect{i}_{\uparrow}$ are color coded. Simulated density: 100$trees/ha$ on a 32x32$m^2$ area.%
    }
    \label{FIG:codedRV}
\end{figure}
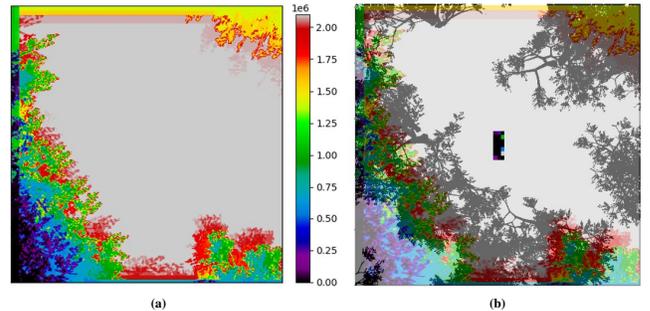
There is however a possibility to encode this information in a way that will allow to extract visibilities of each individual point from the integral $\vect{i}_{\uparrow}$.
The solution we propose here makes the following assumptions: (i) the visibility matrix $\vect{U}_{\uparrow}$ is  binary, and (ii) $K\in\mathbb{N}$ is the number of bits required for representing $\vect{p}_{\uparrow}$, e.g., $K=21$. 
Then, we assign to each ground point of interest in $\vect{p}_{\uparrow}$ a non-binary code \--- a unique positive number $2^k$, $k=\{1,\ldots,K\}$.
Such a number, say $2^4=16$ in a binary representation will have a digit $1$ at one unique position \--- at the $4$th digit in this example.
As such, the binary visibility map associated with this bit can then be ``decoded'' from the integral $\vect{i}_{\uparrow}$.
One can think of these numbers as orthogonal codes in binary domain. 
This will allow de-coupling their contributions from a finite sum $\vect{i}_{\uparrow}$.
Thus, by converting the integral values in $\vect{i}_{\uparrow}$ to a binary representation, one can ``decode'' the visibility of individual ground point of interest. 
A sparse example for such a (coded) integrated visibility map for our $K=21$ ground points of interest is illustrated in Fig. \ref{FIG:codedRV}. 

Note that the coded coefficients in $\vect{i}_{\uparrow}$ must not be normalized.
Although formally $\vect{i}_{\uparrow}$ is no longer a convex combination of columns of $\vect{U}_{\uparrow}$, after decoding the integral visibility it will correspond exactly to a single ground point in $\vect{p}_{\uparrow}$.     
Obviously, the downside of this approach is that the size $K$ of the code words (and thus the number of encodable ground points of interest) depends on the bit-depth of the utilized processing hardware. 
On common CPUs/GPUs, RGB or RGB$\alpha$ images normally have $24$ or $32$ bits per pixel. 
That means that no more than 24 or 32 ground points of interest can be processed in one computation pass. 
If more points need to be supported, multiple computation passes have to be run sequentially for batches of $L$, e.g. being 24 or 32, ground points of interest, with each pass leading to $L$ additional bits of each entry in the integrated visibility map $\vect{i}_{\uparrow}$.
\section{Sampling with RV Constraints} \label{Sec:Sampling}
Given a coded integrated visibility map $\vect{i}_{\uparrow}$ (that represents visibility of ground points of interest $\vect{p}_{\uparrow}$ as seen from the synthetic aperture plane), we can now constrain the sampling at the the synthetic aperture plane to optimize visibility. Considering all possible sampling constellations in $\vect{i}_{\uparrow}$ (i.e., $\sum_{k=1}^{N}k!$ if $\vect{i}_{\uparrow}$ has $N$ entries) to determine a global optimum is combinatorically too complex. A simple greedy choice approximation can find a solution from a given starting position $s$ quickly by maximizing visibility while ensuring sufficient uniformity among all reconstructed $\vect{p}_{\uparrow}$ (Algorithm 1):
\begin{algorithm}
\caption{\textbf{GreedySampling($\vect{i}_{\uparrow},s$)}}\label{greedy_sampling}
\begin{algorithmic}[1]
\State $S=\left\{[s,\vect{i}_{\uparrow}(s)]\right\}$
\While{exit condition not reached}
    \State $\vect{I_a}=BitwiseAverage(S)$
    \For{all $c \in \vect{i}_{\uparrow}$}
        \State  $\vect{I_b}(c) = BitwiseAverage(S \cup \left\{c\right\})$  
    \EndFor
    \State $C = \begin{matrix} \{ c \in \vect{i}_{\uparrow} & | & \left\| \vect{I_b(c)} \right\| > \left\| \vect{I_a} \right\| \} \end{matrix}$ 
    \State $c=Closest(\begin{matrix}  \{ c \in C & | & max\left\| \vect{I_b(c)} - \vect{I_a} \right\| \} \end{matrix},s)$
    \State $S=S \cup \left\{[c,\vect{i}_{\uparrow}(c)]\right\}$
    \State s=c
\EndWhile
\\
\Return ($S$)
\end{algorithmic}
\end{algorithm}
\vspace{0.5mm}
\\
Initially, we store the starting position $s$ and its visibility code $\vect{i}_{\uparrow}(s)$ in sampling set $S$ (line 1). For each iteration, we then compute the bit-wise averages ($\vect{I_a}$ and $\vect{I_b}$) of all codes in $S$ without (line 3) and with (line 5) a possible candidate $c$ in $\vect{i}_{\uparrow}$. We then compute a new set $C$ (line 7) of candidates $c$ for which the visibility increases if considered ($\left\| \vect{I_b(c)} \right\| > \left\| \vect{I_a} \right\|$). From this set, we then select the candidate with maximal visibility gain, and if multiple equal candidates exist, the one closest (shortest distance) to the current sampling position $s$ (line 8). That candidate is finally added to the sampling set $S$ (line 9) and becomes the new current sampling position (line 10). We repeat this procedure until an exit condition is reached (line 2) -- for example if $C$ remains empty for two subsequent iterations, or a maximum number of iterations is completed. 
Note that the bit-wise average of visibility codes represents the results we achieve when computing the integral image by averaging corresponding single images at the selected sample positions, as explained in Sect. \ref{Sec:Introduction}.
Note also that although this method is deterministic for the same start position $s$, the local greedy decision in each iteration does not guarantee to find a global optimal sampling. Therefore we repeat Algorithm 1 for several random start positions and chose the sampling set $S$ with maximal visibility ($max(\left\| \vect{I_a} \right\|)$) and for which the visibility over all $\vect{p}_{\uparrow}$ is sufficiently uniform ($var(\vect{I_a}) < T$, for a given variance threshold $T$). Thus, we do not allow solutions with an extremely high visibility for some points in $\vect{p}_{\uparrow}$ while others have an extremely low visibility.

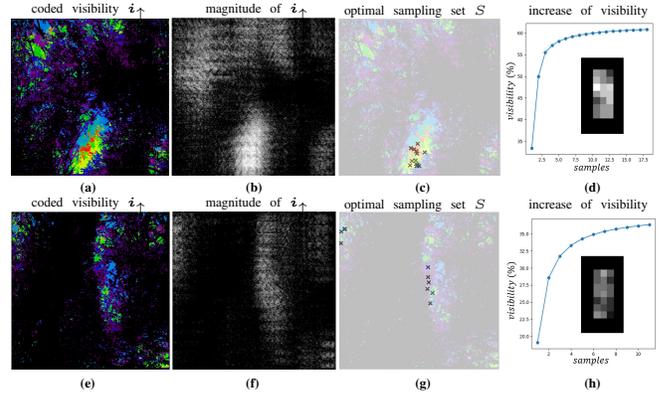
\begin{figure}[!h]
    \centering
    \begin{tikzpicture}
    \node[anchor=south west,inner sep=0] (column) at (-0.4,0) 
    {\resizebox{0.98\linewidth}{!}{\includegraphics{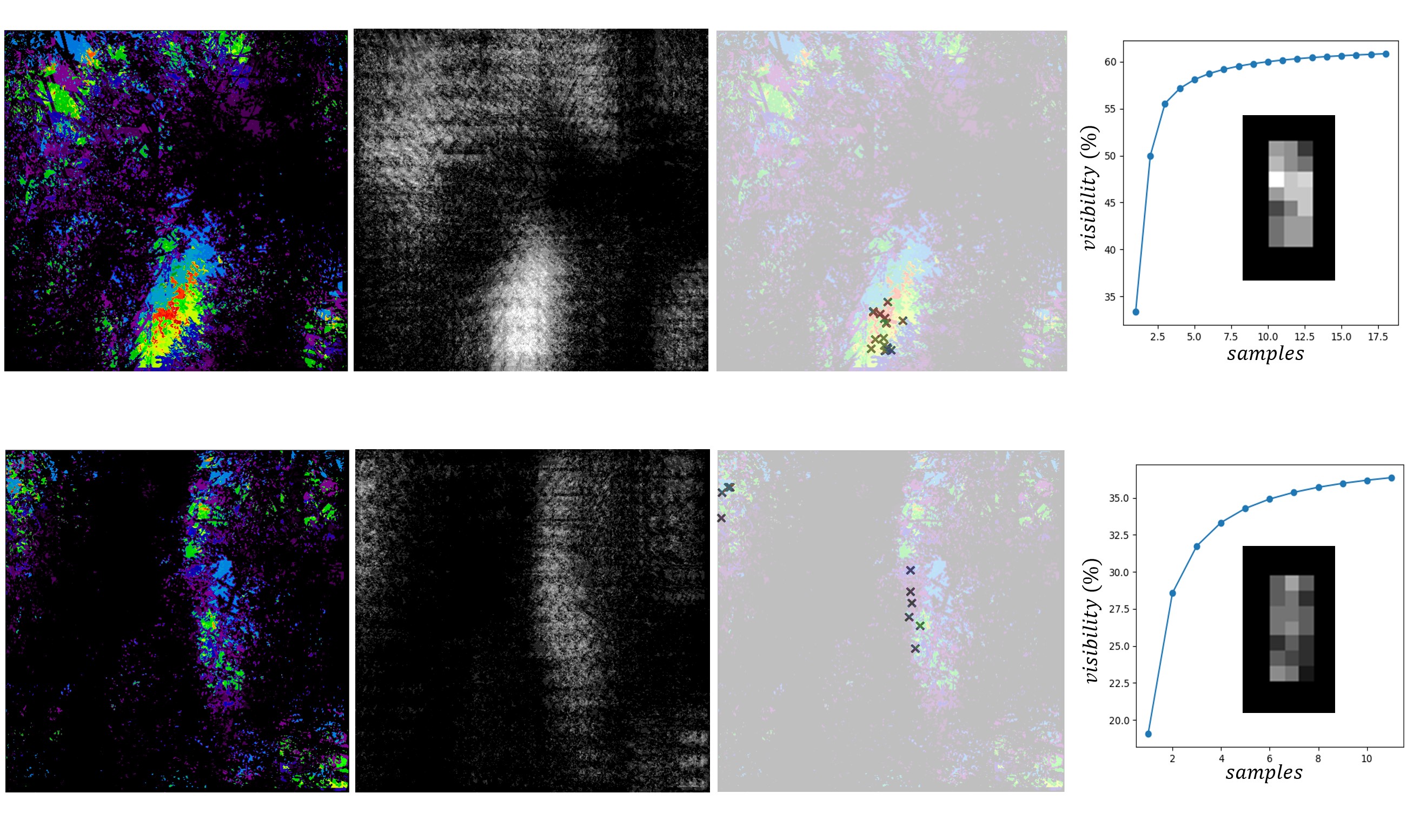}}}; 
    \node [anchor=center,text width=2cm] (eq_col_a) at (0.9,5.0) {
        \begin{minipage}{\textwidth}
        \tiny
        coded visibility $\vect{i}_{\uparrow}$\\
        \end{minipage}
    };
    \node [anchor=center,text width=0.5cm] (eq_col_a) at (0.8,2.7) 
    {{\tiny\textbf{(a)}}};
    
    \node [anchor=center,text width=2cm] (eq_col_a) at (2.9,5.05) {
        \begin{minipage}{\textwidth}
        \tiny\centering
        magnitude of $\vect{i}_{\uparrow}$\\
        \end{minipage}
    };
    \node [anchor=center,text width=0.5cm] (eq_col_a) at (3,2.7) 
    {{\tiny\textbf{(b)}}};
    
    \node [anchor=center,text width=2cm] (eq_col_a) at (5,5.05) {
        \begin{minipage}{\textwidth}
        \tiny\centering
        optimal sampling set $S$\\
        \end{minipage}
    };
    \node [anchor=center,text width=0.5cm] (eq_col_a) at (5.25,2.7) 
    {{\tiny\textbf{(c)}}};

    \node [anchor=center,text width=2cm] (eq_col_a) at (7.3,5.05) {
        \begin{minipage}{\textwidth}
        \tiny\centering
        increase of visibility
        \end{minipage}
    };
    \node [anchor=center,text width=0.5cm] (eq_col_a) at (7.5,2.7) 
    {{\tiny\textbf{(d)}}};

    \node [anchor=center,text width=2cm] (eq_col_a) at (0.9,2.47) {
        \begin{minipage}{\textwidth}
        \tiny
        coded visibility $\vect{i}_{\uparrow}$
        \end{minipage}
    };
    \node [anchor=center,text width=0.5cm] (eq_col_a) at (0.8,0.1) 
    {{\tiny\textbf{(e)}}};
    
    \node [anchor=center,text width=2cm] (eq_col_a) at (2.9,2.47) {
        \begin{minipage}{\textwidth}
        \tiny\centering
        magnitude of $\vect{i}_{\uparrow}$
        \end{minipage}
    };
    \node [anchor=center,text width=0.5cm] (eq_col_a) at (3,0.1) 
    {{\tiny\textbf{(f)}}};
    
    \node [anchor=center,text width=2cm] (eq_col_a) at (5,2.47) {
        \begin{minipage}{\textwidth}
        \tiny\centering
        optimal sampling set $S$
        \end{minipage}
    };
    \node [anchor=center,text width=0.5cm] (eq_col_a) at (5.25,0.1) 
    {{\tiny\textbf{(g)}}};

    \node [anchor=center,text width=2cm] (eq_col_a) at (7.3,2.47) {
        \begin{minipage}{\textwidth}
        \tiny\centering
        increase of visibility
        \end{minipage}
    };
    \node [anchor=center,text width=0.5cm] (eq_col_a) at (7.5,0.1) 
    {{\tiny\textbf{(h)}}};

  \end{tikzpicture}
    \vspace{-0.6cm}
    \caption{%
    Greedy sampling for different forest densities (a-d: 400$trees/ha$, e-h: 500$trees/ha$). Coded integrated visibility maps $\vect{i}_{\uparrow}$ (a,e) and their uncoded magnitudes (b,f) reveal clear patches of better and worse visibility. Note that the ground points of interest $\vect{p}_{\uparrow}$ and their codings are as shown in Fig. \ref{FIG:codedRV}b. The determined set of optimal sampling points $S$ for 50 random start positions and a variance threshold of $T=33\%$ (c,g). Plotted increase of visibility over sequential sampling steps in $S$ (d,h). Here, the inlay illustrates the final reconstruction of the occluded target pattern by integrating the samples $S$. 
    Simulated area for both cases: on 32x32$m^2$.}
    \label{FIG:GREEDY}
\end{figure}

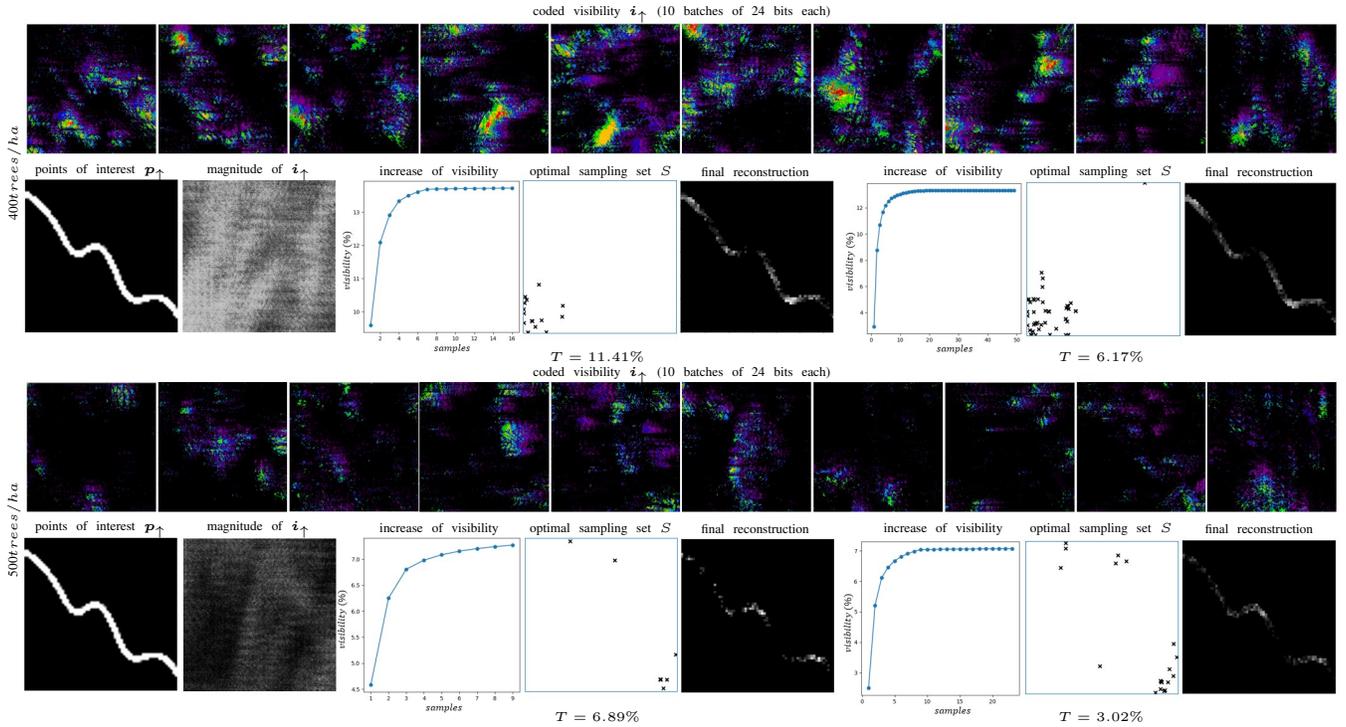
\begin{figure*}[!ht]
    \centering
    \begin{tikzpicture}
    \node[anchor=south west,inner sep=0] (column) at (-0.4,0) 
    {\resizebox{0.98\linewidth}{!}{\includegraphics{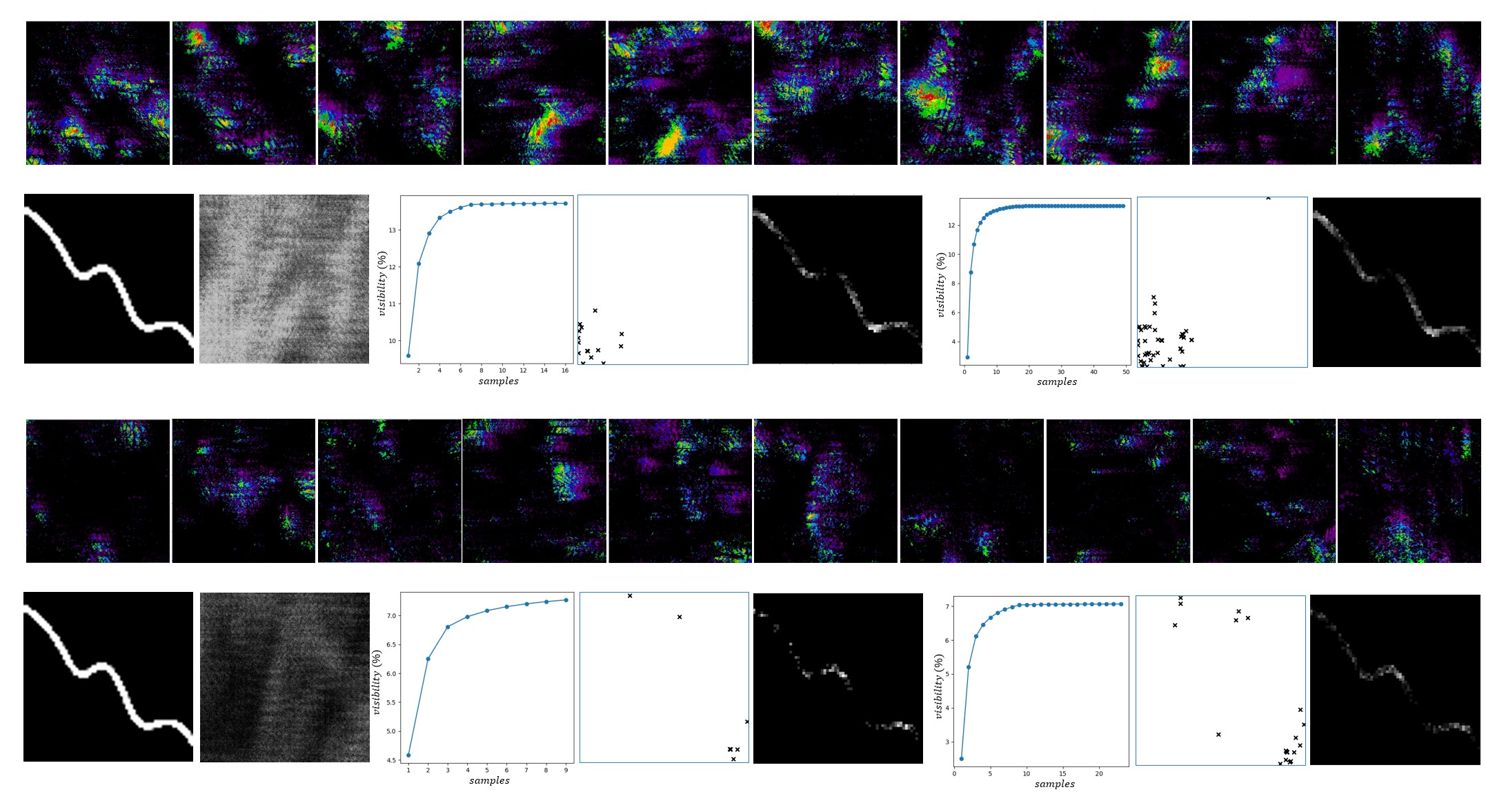}}}; 
    \node [anchor=center,text width=10cm] (eq_col_a) at (11.65,9.55) {
        \begin{minipage}{\textwidth}
        \tiny
        coded visibility $\vect{i}_{\uparrow}$ (10 batches of 24 bits each)\\
        \end{minipage}
    };
    \node [anchor=center,text width=10cm] (eq_col_a) at (11.65,4.75) {
        \begin{minipage}{\textwidth}
        \tiny
        coded visibility $\vect{i}_{\uparrow}$ (10 batches of 24 bits each)\\
        \end{minipage}
    };

    \node [anchor=center,text width=2cm] (eq_col_a) at (3,7.5) {
        \begin{minipage}{\textwidth}
        \tiny\centering
        magnitude of $\vect{i}_{\uparrow}$\\
        \end{minipage}
    };

    \node [anchor=center,text width=2cm] (eq_col_a) at (3,2.75) {
        \begin{minipage}{\textwidth}
        \tiny\centering
        magnitude of $\vect{i}_{\uparrow}$\\
        \end{minipage}
    };

    \node [anchor=center,text width=2cm] (eq_col_a) at (7.55,7.5) {
        \begin{minipage}{\textwidth}
        \tiny\centering
        optimal sampling set $S$\\
        \end{minipage}
    };

    \node [anchor=center,text width=2cm] (eq_col_a) at (14.2,7.5) {
        \begin{minipage}{\textwidth}
        \tiny\centering
        optimal sampling set $S$\\
        \end{minipage}
    };

    \node [anchor=center,text width=2cm] (eq_col_a) at (7.55,2.75) {
        \begin{minipage}{\textwidth}
        \tiny\centering
        optimal sampling set $S$\\
        \end{minipage}
    };

    \node [anchor=center,text width=2cm] (eq_col_a) at (14.2,2.75) {
        \begin{minipage}{\textwidth}
        \tiny\centering
        optimal sampling set $S$\\
        \end{minipage}
    };

    \node [anchor=center,text width=2cm] (eq_col_a) at (5.4,7.5) {
        \begin{minipage}{\textwidth}
        \tiny\centering
        increase of visibility
        \end{minipage}
    };

    \node [anchor=center,text width=2cm] (eq_col_a) at (12.1,7.5) {
        \begin{minipage}{\textwidth}
        \tiny\centering
        increase of visibility
        \end{minipage}
    };
    
    \node [anchor=center,text width=2cm] (eq_col_a) at (5.4,2.75) {
        \begin{minipage}{\textwidth}
        \tiny\centering
        increase of visibility
        \end{minipage}
    };

    \node [anchor=center,text width=2cm] (eq_col_a) at (12.1,2.75) {
        \begin{minipage}{\textwidth}
        \tiny\centering
        increase of visibility
        \end{minipage}
    };

    \node [anchor=center,text width=2cm] (eq_col_a) at (9.6,7.5) {
        \begin{minipage}{\textwidth}
        \tiny\centering
        final reconstruction
        \end{minipage}
    };

    \node [anchor=center,text width=2cm] (eq_col_a) at (16.3,7.5) {
        \begin{minipage}{\textwidth}
        \tiny\centering
        final reconstruction
        \end{minipage}
    };
    
    \node [anchor=center,text width=2cm] (eq_col_a) at (9.6,2.75) {
        \begin{minipage}{\textwidth}
        \tiny\centering
        final reconstruction
        \end{minipage}
    };

    \node [anchor=center,text width=2cm] (eq_col_a) at (16.3,2.75) {
        \begin{minipage}{\textwidth}
        \tiny\centering
        final reconstruction
        \end{minipage}
    };

    \node [anchor=center,text width=2cm] (eq_col_a) at (0.9,7.5) {
        \begin{minipage}{\textwidth}
        \tiny\centering
        points of interest $\vect{p}_{\uparrow}$
        \end{minipage}
    };

    \node [anchor=center,text width=2cm] (eq_col_a) at (0.9,2.75) {
        \begin{minipage}{\textwidth}
        \tiny\centering
        points of interest $\vect{p}_{\uparrow}$
        \end{minipage}
    };

    \node [anchor=center,text width=2cm,rotate=90] (eq_col_a) at (-0.25,7.5) {
        \begin{minipage}{\textwidth}
        \tiny\centering
        400$trees/ha$
        \end{minipage}
    };

    \node [anchor=center,text width=2cm,rotate=90] (eq_col_a) at (-0.25,2.75) {
        \begin{minipage}{\textwidth}
        \tiny\centering
        500$trees/ha$
        \end{minipage}
    };
    
    \node [anchor=center,text width=2cm] (eq_col_a) at (7.5,0.25) {
        \begin{minipage}{\textwidth}
        \tiny\centering
        $T=6.89\%$
        \end{minipage}
    };

    \node [anchor=center,text width=2cm] (eq_col_a) at (14.2,0.25) {
        \begin{minipage}{\textwidth}
        \tiny\centering
        $T=3.02\%$
        \end{minipage}
    };
    
    \node [anchor=center,text width=2cm] (eq_col_a) at (7.5,5.05) {
        \begin{minipage}{\textwidth}
        \tiny\centering
        $T=11.41\%$
        \end{minipage}
    };

    \node [anchor=center,text width=2cm] (eq_col_a) at (14.2,5.05) {
        \begin{minipage}{\textwidth}
        \tiny\centering
        $T=6.17\%$
        \end{minipage}
    };

    \end{tikzpicture}

    \vspace{-0.2cm}
    \caption{%
    Coded integrated visibility map (10 batches of 24 bits each to represent 240 ground points of interest) and corresponding visibility magnitudes (scaled for better visibility) of a forest path covered by 400$trees/ha$ and 500$trees/ha$. A lower variance threshold $T$ leads to more uniform reconstructions with more samples required and slightly less overall visibility (for 400 $trees/ha$: 13.72\% visibility and 16 samples for $T$=11.41\% variance vs. 13.32\% visibility with 49 samples for $T$=6.17\%; for 500 $trees/ha$: 7.27\% visibility with 9 samples for $T$=6.89\% variance vs. 7.07\% visibility with 23 samples for $T$=3.02\%). Simulated area for both cases:  32x32$m^2$. For all cases, 50 random start positions were used.}
    \label{FIG:RESULTS}
     \vspace{-0.2cm}
\end{figure*}

The examples illustrated in Fig. \ref{FIG:GREEDY} show that even with a small number of samples (18 for 400$trees/ha$ and 11 for 500$trees/ha$) relative high target visibilities (60.8\% for 400$trees/ha$ and 36.4\% for 500$trees/ha$) can be achieved. For comparison, the unguided particle swarm optimization (POS) presented in \cite{Nathan23b} achieves, for targets of a similar size, 42\% with 160 samples for 400$trees/ha$ and 31\% with 190 samples for 500$trees/ha$.   
Once all sampling positions are determined, they can be covered either sequentially by a single drone (here, a shortest-path algorithm can bee applied for path-planning \cite{Madkour17}), or in sequential batches of $N$ parallel samples captured by a swarm of $N$ drones (here, closest samples in subsequent batches can be determined with minimum cost bipartite matching \cite{Asathulla19}). 
\\
While in all previous examples $\vect{p}_{\uparrow}$ was a simple rectangular region of interest on the ground, the example in Fig. \ref{FIG:RESULTS} covers the more sophisticated shape of a forest path. Since the path is sampled with 240 points, the 240 bits of its integrated visibility code are computed and represented in 10x 24-bit batches, as explained in Sect. \ref{Sec:Implementation}.    
\section{Conclusion and Outlook}\label{Sec:Conclusion}
In this article we show by simulation, that pre-scanned point-clouds (although not directly suitable for occlusion removal) can guide real-time synthetic aperture sampling. Our approach is scalable with respect to the covered area. For real-time monitored static ground regions and widely unchanged forests, the visibility map is static and can be pre-computed. In future, our principle has to be demonstrated under real conditions in which point-cloud data of LiDAR scans (which are the basis for determining $\vect{i}_{\uparrow}$) is not only sparse but sensor data (e.g., from GPS, IMU, and compass) is also imprecise. Here, an additional stochastic sampling (such as random sampling as done in particle swarm optimization \cite{Nathan23b}) within the proximity of the RV-constrained sampling positions could be added. Furthermore, $\vect{i}_{\uparrow}$ can initially be down-sampled or segmented to ensure visibility clusters of minimal size that better reflect imprecision. The investigation of extensions to better cope with real data is part of our future work. 
\vspace{-0.30cm}
\vspace{0.05cm}
\section*{Acknowledgment}
Thanks to Otto von Guericke University for support in the simulation. This research was funded by the Austrian Science Fund (FWF) and German Research Foundation (DFG) under grant numbers P32185-NBL and I 6046-N, as well as by the State of Upper Austria and the Austrian Federal Ministry of Education, Science and Research via the LIT-Linz Institute of Technology under grant number LIT2019-8-SEE114. 

\ifCLASSOPTIONcaptionsoff
  \newpage
\fi








\vspace{-0.30cm}
\begin{thebibliography}{1}

\bibitem{Kurmi19}
I. Kurmi, D. Schedl, and O. Bimber, Airborne Optical Sectioning,  J. Imaging, vol. 4, no. 102, 2018.

\bibitem{Schedl20b}
D. C. Schedl, I. Kurmi and O. Bimber, Search and rescue with airborne optical sectioning, Nature Machine Intelligence, 2, 783–790, 2020.

\bibitem{Schedl21}
D. C. Schedl, I. Kurmi and O. Bimber,  An Autonomous Drone for Search and Rescue in Forests using Airborne Optical Sectioning,  Science Robotics, 6(55), eabg1188, 2021. 

\bibitem{Schedl20a}
D. C. Schedl, I. Kurmi and O. Bimber, Airborne optical sectioning for nesting observation, Nature Scientific Reports, vol. 10, pp. 7254, 2020.

\bibitem{Youssef23}
M. Youssef and O. Bimber, Fusion of Single and Integral Multispectral Aerial Images, under review, 2023.

\bibitem{reed1990adaptive}
I.S. Reed, and X. Yu, Adaptive multiple-band CFAR detection of an optical pattern with unknown spectral distribution, IEEE transactions on acoustics, speech, and signal processing, vol. 38, no. 10, pp. 1760--1770, 1990. 

\bibitem{Nathan23a}
R. J. A. A. Nathan and O. Bimber, Synthetic Aperture Anomaly Imaging for Through-Foliage Target Detection, Remote Sensing, Volume 15, Number 18, 2023.

\bibitem{Nathan22}
R. J. A. A. Nathan, Indrajit Kurmi, David C. Schedl and Oliver Bimber, Through-Foliage Tracking with Airborne Optical Sectioning, Journal of Remote Sensing, Volume 2022, Article ID 9812765, 2022.

\bibitem{Nathan23b}
R. J. A. A. Nathan, I. Kurmi and O. Bimber, Drone swarm strategy for the detection and tracking of occluded targets in complex environments, Nature Communications Engineering 2, 55, 2023.

\bibitem{Stokes49}
G.G. Stokes, On the perfect blackness of the central spot in Newton's rings, and on the verification of Fresnel's formulae for the intensities of reflected and refracted rays, Cambridge and Dublin Mathematical Journal. new series. 4: 1-14, 1849.

\bibitem{Helmholtz56}
H. Helmholtz, Handbuch der physiologischen Optik, first edition cited by Planck, Leopold Voss, Leipzig, volume 1, page 169, 1856.

\bibitem{Sen05}
P. Sen, B. Chen, G. Garg, S. R. Marschner, M. Horowitz, M. Levoy, H. P. A. Lensch, Dual photography, Proc. ACM SIGGRAPH, pp. 745–755, 2005.

\bibitem{Madkour17}
A. Madkour, W. G. Aref, F. U. Rehman, M. A. Rahman, and S. Basalamah, A survey of shortest-path algorithms. arXiv preprint arXiv:1705.02044, 2017.

\bibitem{Asathulla19}
M. K. Asathulla, S. Khanna, N. Lahn, and S. Raghvendra, A Faster Algorithm for Minimum-cost Bipartite Perfect Matching in Planar Graphs, ACM Transactions on Algorithms, vol. 16, issue  1, article no. 2, pp 1–30, 2019.

\end{thebibliography}
\end{document}